\documentclass[conference]{IEEEtran}
\IEEEoverridecommandlockouts
\usepackage[table,xcdraw]{xcolor}
\usepackage{multirow}
\usepackage{cite}
\usepackage{amsmath,amssymb,amsfonts}
\usepackage{algorithmic}
\usepackage{graphicx}
\usepackage{booktabs}
\usepackage{textcomp}
\usepackage{xcolor}
\usepackage{booktabs}
\def\BibTeX{{\rm B\kern-.05em{\sc i\kern-.025em b}\kern-.08em
    T\kern-.1667em\lower.7ex\hbox{E}\kern-.125emX}}
\begin{document}

\title{Radiomics- and Clinical Feature–Driven Prediction of Volumetric Response in Skull-Base Meningioma after CyberKnife Radiosurgery\\
\thanks{\\
$^{*}$These authors contributed equally as co-last authors.\\
$^{1}$This author is also affiliated with Fondazione IRCCS istituto Neurologico Carlo Besta}
}

\author{
\centering
\IEEEauthorblockN{1\textsuperscript{st} Yin Lin}
\IEEEauthorblockA{\textit{DEIB} \\
\textit{Polytechnic University of Milan}\\
Milan, Italy \\
yin.lin@polimi.it}
\and
\IEEEauthorblockN{2\textsuperscript{nd} Elena De Martin}
\IEEEauthorblockA{\textit{Health Department} \\
\textit{Fondazione IRCCS Istituto Neurologico}\\ \textit{Carlo Besta}\\
Milan, Italy \\
elena.demartin@istituto-besta.it}
\and
\IEEEauthorblockN{3\textsuperscript{rd} Giacomo Conte}
\IEEEauthorblockA{\textit{DEIB} \\
\textit{Polytechnic University of Milan}\\
Milan, Italy \\
giacomo.conte@mail.polimi.it}
\and
\IEEEauthorblockN{4\textsuperscript{th} Domenico Aquino}
\IEEEauthorblockA{\textit{Neuroradiology Unit} \\
\textit{Fondazione IRCCS Istituto Neurologico}\\ \textit{Carlo Besta}\\
Milan, Italy \\
domenico.aquino@istituto-besta.it}
\and
\IEEEauthorblockN{5\textsuperscript{th} Cristiana Pedone}
\IEEEauthorblockA{\textit{Radiotherapy Unit} \\
\textit{Fondazione IRCCS Istituto Neurologico}\\ \textit{Carlo Besta}\\
Milan, Italy\\
cristiana.pedone@istituto-besta.it}
\and
\IEEEauthorblockN{6\textsuperscript{th} Alberto Redaelli}
\IEEEauthorblockA{\textit{DEIB} \\
\textit{Polytechnic University of Milan}\\
Milan, Italy\\
alberto.redaelli@polimi.it}
\and
\IEEEauthorblockN{7\textsuperscript{th} Riccardo Barbieri}
\IEEEauthorblockA{\textit{DEIB} \\
\textit{Polytechnic University of Milan}\\
Milan, Italy \\
riccardo.barbieri@polimi.it}
\and
\IEEEauthorblockN{8\textsuperscript{th} Laura Fariselli$^{*}$}
\IEEEauthorblockA{\textit{Radiotherapy Unit} \\
\textit{Fondazione IRCCS Istituto Neurologico}\\ \textit{Carlo Besta}\\
Milan, Italy \\
laura.fariselli@istituto-besta.it}
\and
\IEEEauthorblockN{9\textsuperscript{th} Simona Ferrante$^{*}$$^{1}$}
\IEEEauthorblockA{\textit{DEIB} \\
\textit{Polytechnic University of Milan}\\
Milan, Italy \\
simona.ferrante@polimi.it}

}

\maketitle

\begin{abstract}
Skull-base meningiomas are often characterized by favorable long-term prognosis, yet their anatomical complexity and proximity to critical neurovascular structures make treatment selection challenging. Stereotactic radiosurgery with CyberKnife represents an effective therapeutic option when surgical resection is not feasible; however, not all patients benefit equally from this treatment. Early identification of patients likely to respond to radiosurgery remains an open clinical problem.
In this study, we propose a radiomics- and clinical feature–driven framework for predicting volumetric response in skull-base meningiomas treated with CyberKnife. Unlike most existing approaches that focus on progression-free survival or recurrence, our method targets volumetric response as an indicator of treatment efficacy. Pre-treatment MRI images from 104 patients were processed to extract radiomic features, which were combined with clinical variables and analyzed using six models. To ensure methodological rigor, the entire modeling process was implemented within a nested cross-validation scheme.
Among the evaluated models, TabPFN achieved the best overall performance, with an AUC of 0.81 and consistently favorable classification metrics. These results suggest that advanced machine learning architectures, when combined with robust validation strategies, can effectively capture patterns associated with treatment response even in small-sample, high-dimensional settings.
\end{abstract}


\section{Introduction}
Although skull base meningiomas do not represent the most common anatomical location overall, they are clinically highly relevant due to their size, anatomical complexity, and growth pattern, which can lead to significant neurological complications, particularly due to the involvement of critical neurovascular structures \cite{meling2019meningiomas,ilyas2019preoperative}.
In symptomatic patients with a marked reduction in quality of life, surgical treatment represents the first-line approach \cite{lemee2019extent}; however, complete resection may not be feasible in cases in which the tumor mass is in close proximity to or encases critical structures . In such situations, radiotherapy and, in particular, stereotactic radiosurgery constitute an effective therapeutic alternative \cite{elias2025efficacy}. Among radiosurgical techniques, CyberKnife has emerged as one of the most effective options due to its high precision of irradiation \cite{colombo2009cyberknife}, its ability to adapt to patient movements, and the possibility of delivering ablative doses while maintaining healthy surrounding tissues. However, not all patients exhibit a favorable response to radiosurgical treatment. In addition, in some cases, radiosurgery may be associated with adverse effects that outweigh the expected benefits. Therefore, the ability to predict treatment outcome prior to irradiation plays a crucial role, enabling better patient selection and optimization of the risk–benefit ratio.
In the literature, many studies have used progression-free survival (PFS) or overall survival (OS) as the main clinical endpoints \cite{lin2025glioblastoma,lin2026lightweight}. A recent meta-analysis \cite{abualnaja2024machine} including 32 studies showed that models integrating clinical and radiomic data outperform those based on a single type of information, achieving AUC values ranging from 0.74 to 0.81 in the prediction of PFS. Similarly, multicenter studies on atypical and grade 2 meningiomas have demonstrated that combined clinico-radiomic models significantly improve recurrence prediction and the identification of high-risk patients who are candidates for adjuvant radiotherapy \cite{ren2024development,park2022interpretable}. Beyond imaging-based approaches, predictive models based on molecular biomarkers, such as the targeted gene expression biomarker described by Chen et al. \cite{chen2023targeted}, as well as models relying exclusively on clinical variables, as reported in the work by Wang et al. \cite{wang2023outcomes}, have also been proposed.
Despite skull meningiomas generally exhibiting a favorable prognosis, with long-term survival rates exceeding 80\% in benign meningiomas \cite{meling2019meningiomas}, the use of traditional targets such as progression-free survival (PFS) has limited clinical relevance for supporting personalized treatment decisions \cite{han2024progression}. In this context, tumor volumetric response following radiotherapy represents a more direct and clinically meaningful target, as it immediately reflects treatment efficacy on tumor burden and may influence patient management. However, this outcome remains surprisingly underexplored in the existing literature.
Initial attempts to address this issue have only recently emerged with the introduction of radiomics-based approaches combined with machine learning. In a pioneering exploratory study, Speckter et al. \cite{speckter2018pretreatment} investigated the correlation between texture features extracted from pre-treatment MRI (T1- and T2-weighted images) and tumor volumetric reduction in an internal cohort of 32 patients. Subsequently, a retrospective study involving 93 patients employed radiomic regression models to predict the monthly volumetric change following Gamma Knife treatment \cite{speckter2022mri}.
Despite their exploratory value, these studies suffer from substantial methodological limitations. In particular, small sample sizes, reliance on relatively simple statistical or regression-based models, suboptimal handling of imbalanced datasets, and the use of weak validation strategies, often limited to a single training–testing split, undermine the generalizability and clinical reliability of the reported findings.

To address these gaps, the present work proposes a novel predictive paradigm centered on volumetric response to Cyberknife as a primary target, more closely aligned with clinical needs. Our study leverages a larger cohort from the Istituto Neurologico Carlo Besta, and systematically integrates radiomic and clinical features, employing a combination of advanced machine learning models and transformer-based architectures within a rigorous nested cross-validation pipeline. This framework is designed to maximize statistical robustness, mitigate overfitting, and enhance predictive generalizability, thereby providing a methodological and clinical contribution beyond the current state of the art.

\section{METHODS}
\begin{figure*}[t] 
    \centering
    \includegraphics[width=18cm]{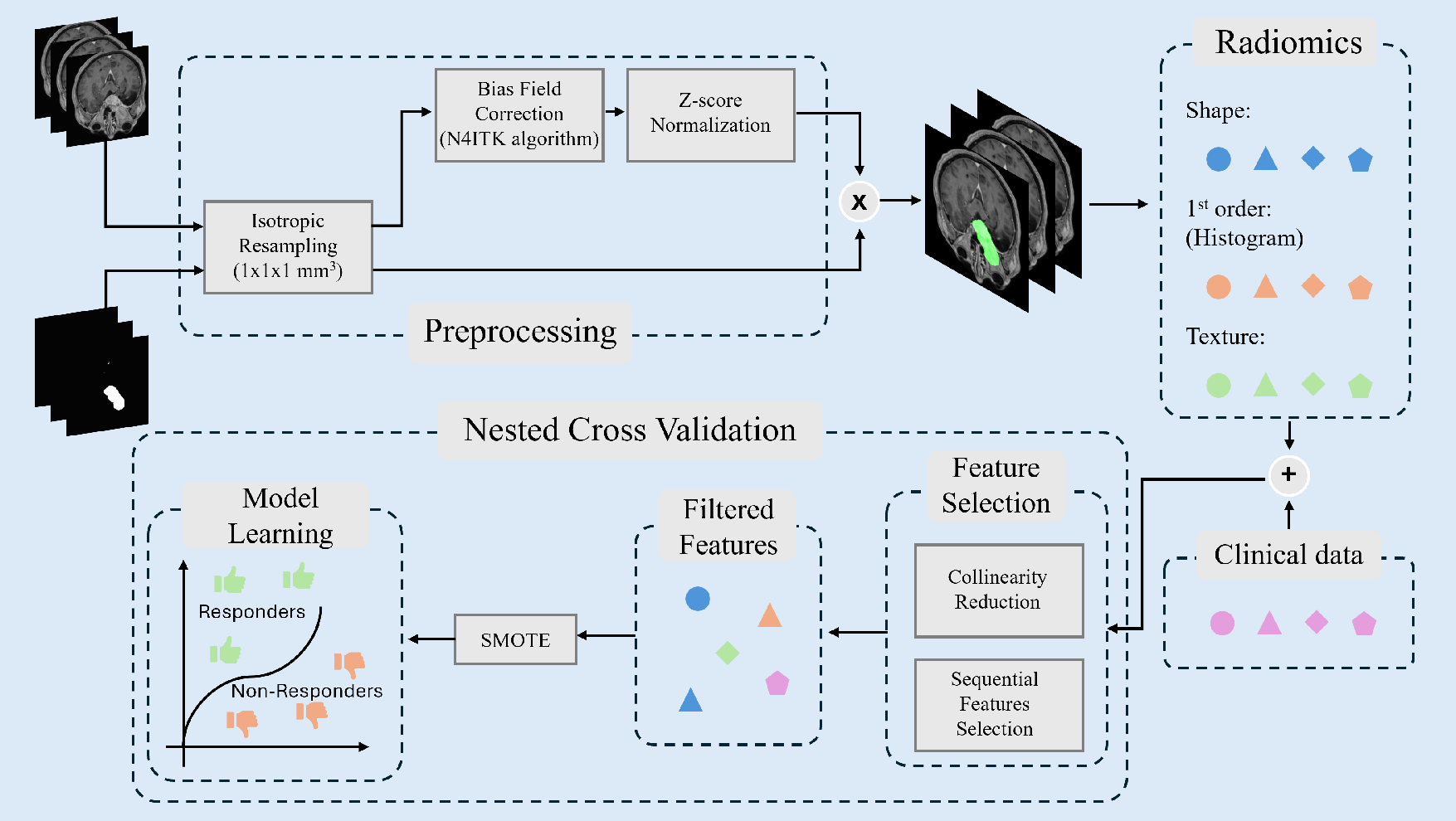}
    \caption{Overview of the proposed radiomics-based pipeline for volumetric response prediction}
    \label{fig:pipeline}
\end{figure*}

The method section of this study is organized as follows. Subsection A describes the dataset and target definition, Subsection B details the MRI preprocessing and radiomic feature extraction procedures, Subsection C and D present the nested cross-validation strategy and feature selection pipeline, and Subsection E outlines the classification models and evaluation metrics. The complete workflow of the proposed approach is schematically summarized in Figure \ref{fig:pipeline}.

\subsection{Data}
The study protocol and the relevant informed consent were approved by the Institutional Ethics Committee (August 11, 2011, approval number 415/2011, Comitato Etico Regione Lombardia, section Fondazione IRCCS Istituto Neurologico). The dataset comprises 104 patients treated with CyberKnife at the Istituto Neurologico Carlo Besta (INCB). The main clinical and treatment characteristics of the cohort are summarized in the table \ref{tab:dataset}. For each patient, the database includes several clinical variables, such as sex, tumor location, baseline tumor volume, age at treatment, and prescription dose. Futhermore, pre-treatment imaging data were available for all patients and consisted of T1-weighted magnetic resonance imaging (MRI) scans fused with computed tomography (CT) images. Tumor segmentations were provided for each image volume and were performed by radiologists at INCB using semi-automatic software routinely employed for radiotherapy treatment planning. The images had in-plane dimensions of 350 mm × 350 mm, while the extent along the z-axis (number of longitudinal slices) varied across patients, with a median value of 330 slices. 

In addition, a predictive target variable was defined. Volumetric response was defined as the percentage change in tumor volume at the final follow-up, conducted no earlier than 36 months after initial radiosurgery, relative to baseline (pre-treatment) volume. This outcome was categorized as follows: Partial Response (PR), defined as a volume reduction greater than 20\%; Stable Disease (SD), indicating no significant change in volume; and Progressive Disease (PD), defined as a volume increase greater than 20\%.
The target variable was binarized: Stable Disease (SD) and Progressive Disease (PD) were grouped as Non-Responders (Class 0), whereas Partial Response (PR) was defined as Responders (Class 1).

\begin{table}[t]
\caption{Clinical and treatment characteristics of the study cohort}
\label{tab:dataset}
\centering
\resizebox{\columnwidth}{!}{%
\begin{tabular}{|l|l|c|}
\hline
\rowcolor[HTML]{EFEFEF}
\multicolumn{2}{|c|}{\textbf{Characteristics}} & \textbf{Median (range) or number (\%)} \\ \hline

\multicolumn{2}{|l|}{Age at treatment} 
& 57.1 (19.8--86.8) \\ \hline

\multicolumn{2}{|l|}{Sex (female)} 
& 84 (80.8\%) \\ \hline

\multicolumn{2}{|l|}{Tumour volume before treatment (cc)} 
& 11.6 (0.8--57.4) \\ \hline

\multicolumn{2}{|l|}{Mean dose (Gy)} 
& 27.5 (21.9--30.8) \\ \hline

\multirow{3}{*}{Volumetric response} 
& Progressive Response (PR) & 67 (64.4\%) \\ \cline{2-3}
& Stable Disease (SD) & 32 (30.8\%) \\ \cline{2-3}
& Progressive Disease (PD) & 5 (4.8\%) \\ \hline

\end{tabular}%
}
\end{table}

\subsection{Mri Preprocessing and Features Extraction}

In accordance with the guidelines reported in \cite{seoni2024all}, a standard preprocessing pipeline was applied to all imaging data prior to radiomic analysis (Figure \ref{fig:pipeline}). All images were first resampled to an isotropic voxel spacing of $1 \times 1 \times 1 \text{mm}^3$ in order to ensure spatial consistency across the dataset. Subsequently, intensity non-uniformities inherent to MRI acquisitions were corrected using the N4ITK algorithm \cite{tustison2010n4itk}, which is widely recognized as a robust method for mitigating bias field effects caused by magnetic field inhomogeneities. Voxel intensities were then normalized through Z-score standardization, defined as $z = (x - \mu)/\sigma$, where $x$ denotes the original voxel intensity, and $\mu$ and $\sigma$ represent the mean and standard deviation, respectively. To avoid data leakage, normalization was performed independently for each image volume rather than using global dataset statistics. These preprocessing steps were selected to harmonize the imaging data and improve the robustness and reproducibility of the subsequent radiomic analysis. Following preprocessing, radiomic features were extracted using the PyRadiomics library \cite{van2017computational} in a fully three-dimensional (3D) manner. To reduce computational complexity and exclude non-informative regions, feature extraction was restricted to image slices containing tumor segmentations. In order to enrich the feature space, Laplacian of Gaussian (LoG) and wavelet filters were applied to the original images, generating multiple derived image representations and enabling the characterization of tumor properties across different frequency domains. The extracted features comprised shape descriptors capturing the three-dimensional geometry and volumetric properties of the lesions, first-order statistics describing the distribution of voxel intensities, and texture features quantifying spatial relationships among voxels, which are commonly interpreted as markers of intra-tumoral heterogeneity.

\subsection{Nested Cross-Validation Strategy}
To compensate for the absence of an external validation cohort, the machine learning pipeline was implemented using a nested cross-validation (NCV) strategy. This framework enables a reliable assessment of model generalization by reducing the bias introduced by random data partitioning \cite{varma2006bias} and by limiting selection bias during performance evaluation \cite{cawley2010over}. The NCV design comprised two hierarchical levels. In the outer loop, the dataset was divided into five folds, with each fold iteratively used as an independent test set while the remaining folds were used for model training. Within each outer training set, an additional five-fold cross-validation was conducted as an inner loop dedicated exclusively to model selection, including the feature selection procedure and SMOTE. This hierarchical separation ensured that all feature selection steps were performed solely on training data, thereby preventing information leakage. Model performance was ultimately quantified by averaging the results obtained across all outer folds, yielding a robust and unbiased estimate of predictive performance that approximates external validation through repeated testing on unseen data subsets.

\subsection{Feature Selection and Unbalancing Management}
Following radiomic feature extraction, the resulting dataset consisted of 1,032 features derived from 104 patients. Given the pronounced imbalance between feature dimensionality and sample size, the adoption of a feature reduction strategy was essential to mitigate overfitting and enhance model generalizability. Within each outer cross-validation split, the data were partitioned into training (70\%) and test (30\%) subsets using stratified sampling to preserve class distributions. All feature selection procedures and SMOTE were applied exclusively to the training data to prevent information leakage.
Feature selection was implemented through a two-stage pipeline. In the first stage, a univariate statistical screening was conducted to assess the discriminative ability of each feature with respect to the target classes. Continuous variables were evaluated using the Mann–Whitney U test (Mann \& Whitney, 1947), while categorical variables were assessed using Pearson’s chi-squared test (Pearson, 1900). Each feature was associated with a p-value reflecting its statistical relevance. To further reduce redundancy, pairwise Spearman’s rank correlation coefficients were computed among all features. For feature pairs exhibiting high collinearity (correlation coefficient $>$ 0.6), the feature with the weaker statistical association (higher p-value) was discarded.
In the second stage, a wrapper-based Sequential Forward Selection (SFS) \cite{jain2002feature} approach was applied to identify the most informative feature subset. The procedure was initialized by training individual models for each candidate feature and selecting the feature yielding the best performance under internal cross-validation. Subsequently, features were added iteratively in a forward manner, with each step selecting the feature that maximized predictive performance when combined with the previously selected subset. The process was terminated upon reaching a predefined maximum number of features based on the sample size, to avoid issues related to the curse of dimensionality \cite{klontzas2025sample}. The final radiomic signature was defined as the feature subset achieving the optimal predictive performance during internal cross-validation.
To mitigate class imbalance (64\% Responders vs. 36\% Non-Responders), SMOTE was applied within the nested cross-validation framework, restricted to the inner-loop training folds to prevent information leakage \cite{chawla2002smote}.

\subsection{Classification Models and Evaluation Metrics}
A comparative evaluation was performed across a heterogeneous set of classification algorithms in order to identify the most effective modeling strategy. The analysis included a spectrum of approaches spanning classical, ensemble-based, boosting, and transformer-based methods. Specifically, classical classifiers such as the Ridge Classifier and Decision Trees were considered, given their widespread use in radiomics applications \cite{saroh2025machine}. Ensemble learning was represented by the Random Forest algorithm \cite{breiman2001random}. In addition, gradient boosting frameworks, namely XGBoost and CatBoost, were included due to their demonstrated effectiveness in handling complex, high-dimensional medical datasets \cite{chen2016xgboost, prokhorenkova2018catboost}. Finally, a transformer-based architecture, TabPFN, was evaluated as a state-of-the-art Prior-Data Fitted Network pre-trained on large-scale tabular datasets, enabling strong performance in low-sample-size settings \cite{hollmann2025accurate}
Model performance was assessed using multiple complementary evaluation metrics, including the area under the receiver operating characteristic curve (AUC), accuracy, precision, recall, and F1-score, in order to provide a comprehensive characterization of classification performance.

\section{RESULTS}

\begin{table*}[ht]
\centering
\caption{Performance comparison of the evaluated classification models.}
\label{tab:model_performance}
\begin{tabular}{lcccccccccc}
\toprule
\textbf{Model} 
& \multicolumn{2}{c}{\textbf{AUC}} 
& \multicolumn{2}{c}{\textbf{Accuracy}} 
& \multicolumn{2}{c}{\textbf{Recall}} 
& \multicolumn{2}{c}{\textbf{Precision}} 
& \multicolumn{2}{c}{\textbf{F1-score}} \\
\cmidrule(lr){2-3} \cmidrule(lr){4-5} \cmidrule(lr){6-7} \cmidrule(lr){8-9} \cmidrule(lr){10-11}
 & Mean $\uparrow$& Std Dev $\downarrow$ & Mean $\uparrow$& Std Dev $\downarrow$ & Mean $\uparrow$& Std Dev $\downarrow$ & Mean $\uparrow$& Std Dev $\downarrow$ & Mean $\uparrow$& Std Dev $\downarrow$ \\
\midrule
Decision Tree    
& 0.67 & 0.06 & 0.66 & 0.07 & 0.64 & 0.10 & 0.67 & 0.04 & 0.66 & 0.07 \\
Ridge Classifier 
& 0.70 & 0.06 & 0.69 & 0.07 & 0.65 & 0.10 & 0.71 & 0.08 & 0.68 & 0.09 \\
XGBoost          
& 0.72 & 0.07 & 0.69 & 0.08 & 0.70 & 0.06 & 0.74 & 0.04 & 0.72 & 0.05 \\
CatBoost         
& 0.73 & 0.06 & 0.71 & 0.04 & 0.65 & 0.08 & 0.74 & 0.05 & 0.69 & 0.07 \\
Random Forest    
& 0.76 & 0.07 & 0.69 & 0.10 & 0.73 & 0.11 & 0.79 & 0.10 & 0.76 & 0.11 \\
\textbf{TabPFN}  
& \textbf{0.81} & \textbf{0.07} & \textbf{0.76} & \textbf{0.04} & \textbf{0.76} & \textbf{0.08} & \textbf{0.86} & \textbf{0.05} & \textbf{0.81} & \textbf{0.06} \\
\bottomrule
\end{tabular}
\end{table*}

\begin{figure*}[t] 
    \centering
    \includegraphics[width=18cm]{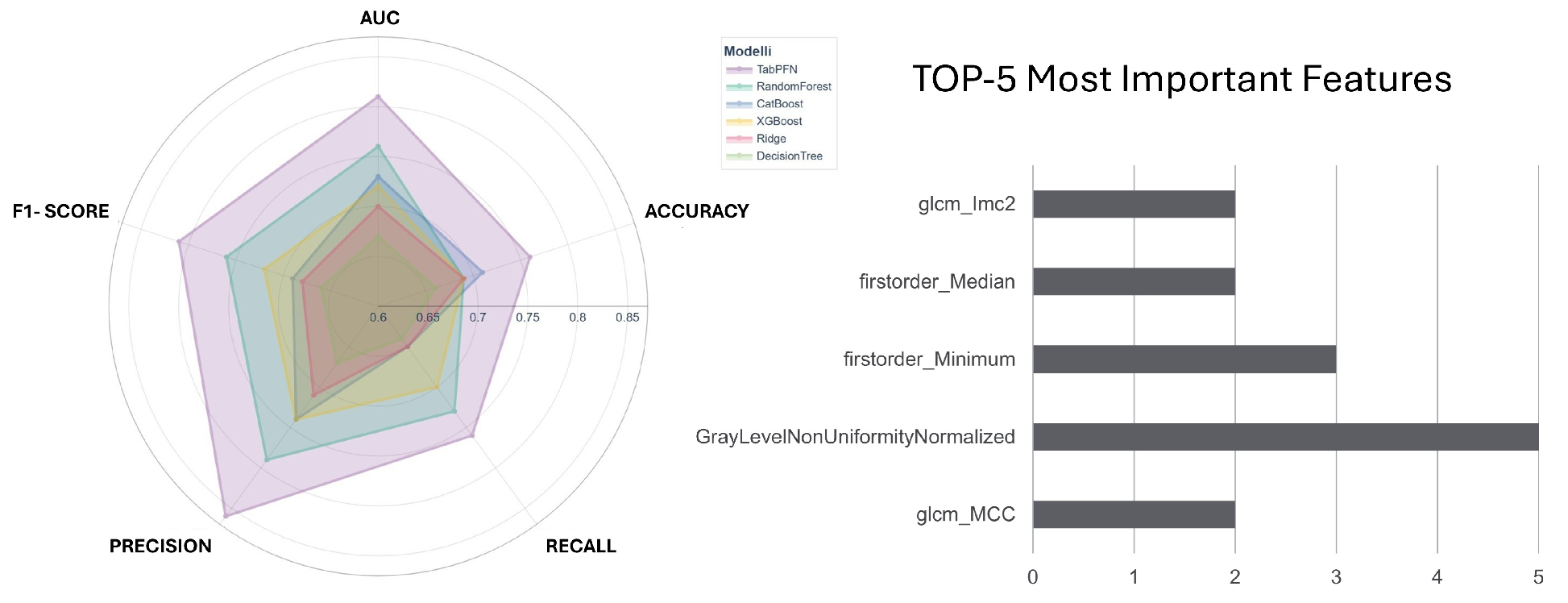}
    \caption{Performance of the evaluated classification models across multiple metrics, along with the five most frequently selected features across the outer cross-validation folds.}
    \label{fig:performance}
\end{figure*}
Across nested cross-validation folds, 11–15 features were selected (median $\approx$13). A stable radiomic core was observed, with $GrayLevelNonUniformityNormalized$ consistently selected in all seeds (Figure \ref{fig:performance}, representing the only feature common to all models. Clinical variables showed limited contribution, with only age and prescription dose intermittently selected.

Classification performance was evaluated using a nested cross-validation (NCV) framework, with metrics computed on the test sets of the outer loop. The quantitative performance of all evaluated classification models is reported in Table \ref{tab:model_performance}, and the comparative performance of all models and the distribution of the top five metrics are illustrated graphically in Figure \ref{fig:performance}. From results, the TabPFN model consistently outperformed all other architectures across every evaluation metric, achieving the highest discriminative capability with an AUC of $0.81 \pm 0.07$. In addition, TabPFN obtained an accuracy of $0.76 \pm 0.04$, recall of $0.76 \pm 0.08$, precision of $0.86 \pm 0.05$, and the highest F1-score of $0.81 \pm 0.06$, while also exhibiting relatively low variability across nested cross-validation folds.


\section{DISCUSSION}

This study introduces an innovative framework based on the integration of radiomic and clinical features for predicting volumetric response in skull-base meningiomas treated with CyberKnife radiosurgery. Unlike most previous works, which have focused on progression-free survival or recurrence, the proposed approach uses volumetric response as the primary outcome, a clinically more direct indicator of treatment efficacy. The combined use of machine learning models and transformer-based architectures, implemented within a nested cross-validation pipeline, enabled a methodologically rigorous validation even in the presence of a limited-size dataset.

From a clinical perspective, the ability to identify patients who are likely to be responsive or non-responsive to treatment prior to irradiation could support therapeutic decision-making, thereby avoiding exposure to potentially unnecessary treatments or treatments associated with unjustified risks, especially in cases of tumors located in proximity to critical structures.

Some limitations should nevertheless be considered. First, the study is based on a single-center retrospective cohort and on a still limited sample size, factors that may limit the generalizability of the results, despite the adoption of nested cross-validation to reduce the risk of overfitting. In addition, the present work does not include an in-depth analysis of explainability, an aspect that is relevant for the clinical interpretation of predictive models.

From a future perspective, the goal will be to expand the cohort through the inclusion of new patients and, where possible, to integrate an external cohort for independent validation. Further developments will include conducting model interpretability analyses aimed at providing a clearer understanding of the clinical, anatomical, or pathological meaning of the most predictive features, thereby facilitating a potential integration of the model into clinical practice.

\section*{Acknowledgment}

This work was supported by the project Cal.Hub.Ria (project code T4-AN-09) funded by the Italian Ministry of Health in the framework of  "Piano Sviluppo e Coesione Salute, FSC 2014-2020".


\begin{thebibliography}{00}
\bibitem{meling2019meningiomas}Meling, T., Da Broi, M., Scheie, D. \& Helseth, E. Meningiomas: skull base versus non-skull base. {\em Neurosurgical Review}. \textbf{42}, 163-173 (2019)

\bibitem{ilyas2019preoperative}Ilyas, A., Przybylowski, C., Chen, C., Ding, D., Foreman, P., Buell, T., Taylor, D., Kalani, M. \& Park, M. Preoperative embolization of skull base meningiomas: a systematic review. {\em Journal Of Clinical Neuroscience}. \textbf{59} pp. 259-264 (2019)


\bibitem{elias2025efficacy}El{\'i}as, J., Cacho, A., Luj{\'a}n, A., L{\'o}pez, J. \& Trejo, J. Efficacy and Safety of Stereotactic Radiosurgery in Patients With Large-Volume Meningiomas $\geq10 cm^{3}$: A Systematic Review and Single-Arm Meta-Analysis. {\em Cureus}. \textbf{17} (2025)

\bibitem{lemee2019extent}Lemée, J., Corniola, M., Da Broi, M., Joswig, H., Scheie, D., Schaller, K., Helseth, E. \& Meling, T. Extent of resection in meningioma: predictive factors and clinical implications. {\em Scientific Reports}. \textbf{9}, 5944 (2019)

\bibitem{lin2025glioblastoma}Lin, Y., Barbieri, R., Aquino, D., Lauria, G., Grisoli, M., De Momi, E., Redaelli, A. \& Ferrante, S. Glioblastoma Overall Survival Prediction With Vision Transformers. {\em 2025 47th Annual International Conference Of The IEEE Engineering In Medicine And Biology Society (EMBC)}. pp. 1-4 (2025)

\bibitem{lin2026lightweight}Lin, Y., Aquino, D., Lauria, G., Grisoli, M., Redaelli, A., Barbieri, R. \& Ferrante, S. Lightweight ensemble vision transformer framework for non-invasive survival prediction in glioblastoma. {\em Neurocomputing}. pp. 133303 (2026)

\bibitem{colombo2009cyberknife}Colombo, F., Casentini, L., Cavedon, C., Scalchi, P., Cora, S. \& Francescon, P. Cyberknife radiosurgery for benign meningiomas: short-term results in 199 patients. {\em Neurosurgery}. \textbf{64}, A7-A13 (2009)

\bibitem{abualnaja2024machine}Abualnaja, S., Morris, J., Rashid, H., Cook, W. \& Helmy, A. Machine learning for predicting post-operative outcomes in meningiomas: a systematic review and meta-analysis. {\em Acta Neurochirurgica}. \textbf{166}, 505 (2024)

\bibitem{ren2024development}Ren, L., Chen, J., Deng, J., Qing, X., Cheng, H., Wang, D., Ji, J., Chen, H., Juratli, T., Wakimoto, H. \& Others The development of a combined clinico-radiomics model for predicting post-operative recurrence in atypical meningiomas: a multicenter study. {\em Journal Of Neuro-Oncology}. \textbf{166}, 59-71 (2024)

\bibitem{park2022interpretable}Park, C., Choi, S., Eom, J., Byun, H., Ahn, S., Chang, J., Kim, S., Lee, S., Park, Y. \& Yoon, H. An interpretable radiomics model to select patients for radiotherapy after surgery for WHO grade 2 meningiomas. {\em Radiation Oncology}. \textbf{17}, 147 (2022)

\bibitem{chen2023targeted}Chen, W., Choudhury, A., Youngblood, M., Polley, M., Lucas, C., Mirchia, K., Maas, S., Suwala, A., Won, M., Bayley, J. \& Others Targeted gene expression profiling predicts meningioma outcomes and radiotherapy responses. {\em Nature Medicine}. \textbf{29}, 3067-3076 (2023)

\bibitem{wang2023outcomes}Wang, J., Landry, A., Nassiri, F., Merali, Z., Patel, Z., Lee, G., Rogers, L., Zuccato, J., Voisin, M., Munoz, D. \& Others Outcomes and predictors of response to fractionated radiotherapy as primary treatment for intracranial meningiomas. {\em Clinical And Translational Radiation Oncology}. \textbf{41} pp. 100631 (2023)

\bibitem{han2024progression}Han, T., Liu, X. \& Zhou, J. Progression/recurrence of meningioma: an imaging review based on magnetic resonance imaging. {\em World Neurosurgery}. \textbf{186} pp. 98-107 (2024)

\bibitem{speckter2018pretreatment}Speckter, H., Bido, J., Hernandez, G., Rivera, D., Suazo, L., Valenzuela, S., Miches, I., Oviedo, J., Gonzalez, C. \& Stoeter, P. Pretreatment texture analysis of routine MR images and shape analysis of the diffusion tensor for prediction of volumetric response after radiosurgery for meningioma. {\em Journal Of Neurosurgery}. \textbf{129}, 31-37 (2018)

\bibitem{speckter2022mri}Speckter, H., Radulovic, M., Trivodaliev, K., Vranes, V., Joaquin, J., Hernandez, W., Mota, A., Bido, J., Hernandez, G., Rivera, D. \& Others MRI radiomics in the prediction of the volumetric response in meningiomas after gamma knife radiosurgery. {\em Journal Of Neuro-Oncology}. \textbf{159}, 281-291 (2022)

\bibitem{seoni2024all}Seoni, S., Shahini, A., Meiburger, K., Marzola, F., Rotunno, G., Acharya, U., Molinari, F. \& Salvi, M. All you need is data preparation: A systematic review of image harmonization techniques in Multi-center/device studies for medical support systems. {\em Computer Methods And Programs In Biomedicine}. \textbf{250} pp. 108200 (2024)

\bibitem{tustison2010n4itk}Tustison, N., Avants, B., Cook, P., Zheng, Y., Egan, A., Yushkevich, P. \& Gee, J. N4ITK: improved N3 bias correction. {\em IEEE Transactions On Medical Imaging}. \textbf{29}, 1310-1320 (2010)

\bibitem{van2017computational}Van Griethuysen, J., Fedorov, A., Parmar, C., Hosny, A., Aucoin, N., Narayan, V., Beets-Tan, R., Fillion-Robin, J., Pieper, S. \& Aerts, H. Computational radiomics system to decode the radiographic phenotype. {\em Cancer Research}. \textbf{77}, e104-e107 (2017)

\bibitem{varma2006bias}Varma, S. \& Simon, R. Bias in error estimation when using cross-validation for model selection. {\em BMC Bioinformatics}. \textbf{7}, 91 (2006)

\bibitem{cawley2010over}Cawley, G. \& Talbot, N. On over-fitting in model selection and subsequent selection bias in performance evaluation. {\em The Journal Of Machine Learning Research}. \textbf{11} pp. 2079-2107 (2010)

\bibitem{jain2002feature}Jain, A. \& Zongker, D. Feature selection: Evaluation, application, and small sample performance. {\em IEEE Transactions On Pattern Analysis And Machine Intelligence}. \textbf{19}, 153-158 (2002)

\bibitem{saroh2025machine}Saroh, S., Pendem, S., Prakashini, K., Nayak, S., Menon, G., Divya, B. \& Others Machine learning based radiomics approach for outcome prediction of meningioma–a systematic review. {\em F1000Research}. \textbf{14} pp. 330 (2025)

\bibitem{breiman2001random}Breiman, L. Random forests. {\em Machine Learning}. \textbf{45}, 5-32 (2001)

\bibitem{chen2016xgboost}Chen, T. XGBoost: A Scalable Tree Boosting System. {\em Cornell University}. (2016)

\bibitem{prokhorenkova2018catboost}Prokhorenkova, L., Gusev, G., Vorobev, A., Dorogush, A. \& Gulin, A. CatBoost: unbiased boosting with categorical features. {\em Advances In Neural Information Processing Systems}. \textbf{31} (2018)

\bibitem{hollmann2025accurate}Hollmann, N., Müller, S., Purucker, L., Krishnakumar, A., Körfer, M., Hoo, S., Schirrmeister, R. \& Hutter, F. Accurate predictions on small data with a tabular foundation model. {\em Nature}. \textbf{637}, 319-326 (2025)

\bibitem{chawla2002smote}Chawla, N., Bowyer, K., Hall, L. \& Kegelmeyer, W. SMOTE: synthetic minority over-sampling technique. {\em Journal Of Artificial Intelligence Research}. \textbf{16} pp. 321-357 (2002)

\bibitem{klontzas2025sample}Klontzas, M., Kocak, B. \& Cuocolo, R. Sample size estimation for radiomics studies: an overlooked problem. {\em European Radiology}. pp. 1-2 (2025)

\end{thebibliography}
\end{document}